\def\paperID{8035}
\def\confName{ICCV}
\def\confYear{2025}

\def\paperTitle{InterSyn: Interleaved Learning for Dynamic Motion Synthesis in the Wild}

\def\authorBlock{
    Yiyi Ma \textsuperscript{1}\thanks{This work was performed at TeleAI.} \qquad 
    Yuanzhi Liang \textsuperscript{2}\qquad
    Xiu Li \textsuperscript{1}\footnotemark[2]\qquad
    Chi Zhang \textsuperscript{2}\qquad
    Xuelong Li \textsuperscript{2}\footnotemark[2]\thanks{Corresponding authors.}\\
    
    Shenzhen International Graduate School, Tsinghua University\textsuperscript{1} \\
    Institute of Artificial Intelligence, China Telecom\textsuperscript{2}
}

\newif\ifreview 
\newif\ifarxiv 
\newif\ifcamera \newcommand{\cameraready}{\cameratrue}
\newif\ifrebuttal 

\cameraready

\pdfoutput=1
\documentclass[10pt,twocolumn,letterpaper]{article}
\ifreview \usepackage[review]{cvpr} \fi
\ifarxiv \usepackage[pagenumbers]{cvpr} \fi
\ifrebuttal \usepackage[rebuttal]{cvpr} \fi
\ifcamera \usepackage{cvpr} \fi

\usepackage{graphicx}	
\usepackage{amsmath}	
\usepackage{amssymb}	
\usepackage{booktabs}
\usepackage{times}
\usepackage{microtype}
\usepackage{epsfig}
\usepackage{caption}
\usepackage{float}
\usepackage{placeins}
\usepackage{color, colortbl}
\usepackage{stfloats}
\usepackage{enumitem}
\usepackage{tabularx}
\usepackage{xstring}
\usepackage{multirow}
\usepackage{xspace}
\usepackage{url}
\usepackage{subcaption}
\usepackage{xcolor}
\usepackage[hang,flushmargin]{footmisc}

\ifcamera \usepackage[accsupp]{axessibility} \fi

% ...

\ifarxiv  \fi

\newcommand{\R}[1]{{%
    \textbf{%
        \ifstrequal{#1}{1}{\textcolor{red}{R#1}}{%
        \ifstrequal{#1}{2}{\textcolor{blue}{R#1}}{%
        \ifstrequal{#1}{3}{\textcolor{magenta}{R#1}}{%
        \ifstrequal{#1}{4}{\textcolor{teal}{R#1}}{%
                           \textcolor{cyan}{R#1}%
        }}}}%
    }%
}}
\usepackage{xr-hyper}

\makeatletter
\newcommand*{\addFileDependency}[1]{
  \typeout{(#1)}
  \@addtofilelist{#1}
  \IfFileExists{#1}{}{\typeout{No file #1.}}
}

\makeatother

\definecolor{cvprblue}{rgb}{0.21,0.49,0.74}
\usepackage[pagebackref,breaklinks,colorlinks,allcolors=cvprblue]{hyperref}
\usepackage[capitalize]{cleveref}
\crefname{section}{Sec.}{Secs.}
\crefname{table}{Table}{Tables}
\crefname{figure}{Fig.}{Figs.}

\ifarxiv \crefname{appendix}{App.}{Apps.}
\else \crefname{appendix}{Suppl.}{Suppls.} \fi

\begin{document}
% %% TITLE

\title{\paperTitle}
\author{\authorBlock}
\maketitle

\begin{abstract}
We present Interleaved Learning for Motion Synthesis (InterSyn), a novel framework that targets the generation of realistic interaction motions by learning from integrated motions that consider both solo and multi-person dynamics. Unlike previous methods that treat these components separately, InterSyn employs an interleaved learning strategy to capture the natural, dynamic interactions and nuanced coordination inherent in real-world scenarios. Our framework comprises two key modules: the Interleaved Interaction Synthesis (INS) module, which jointly models solo and interactive behaviors in a unified paradigm from a first-person perspective to support multiple character interactions, and the Relative Coordination Refinement (REC) module, which refines mutual dynamics and ensures synchronized motions among characters. Experimental results show that the motion sequences generated by InterSyn exhibit higher text-to-motion alignment and improved diversity compared with recent methods, setting a new benchmark for robust and natural motion synthesis. Additionally, our code will be open-sourced in the future to promote further research and development in this area. Project website: \href{https://myy888.github.io/InterSyn/}{https://myy888.github.io/InterSyn/}

% We compared InterSyn with FreeMotion \cite{fan2024freemotion} under the number-free human generation task and found that InterSyn outperforms Free Motion by 40\% in terms of the FID metric. Additionally, InterSyn significantly outperforms Free Motion across all other evaluation metrics, setting a new benchmark for robust and natural motion synthesis.
\end{abstract}

\section{Introduction}
\label{sec:intro}

\begin{figure}[t!]
    \centering
    \includegraphics[width=\linewidth]{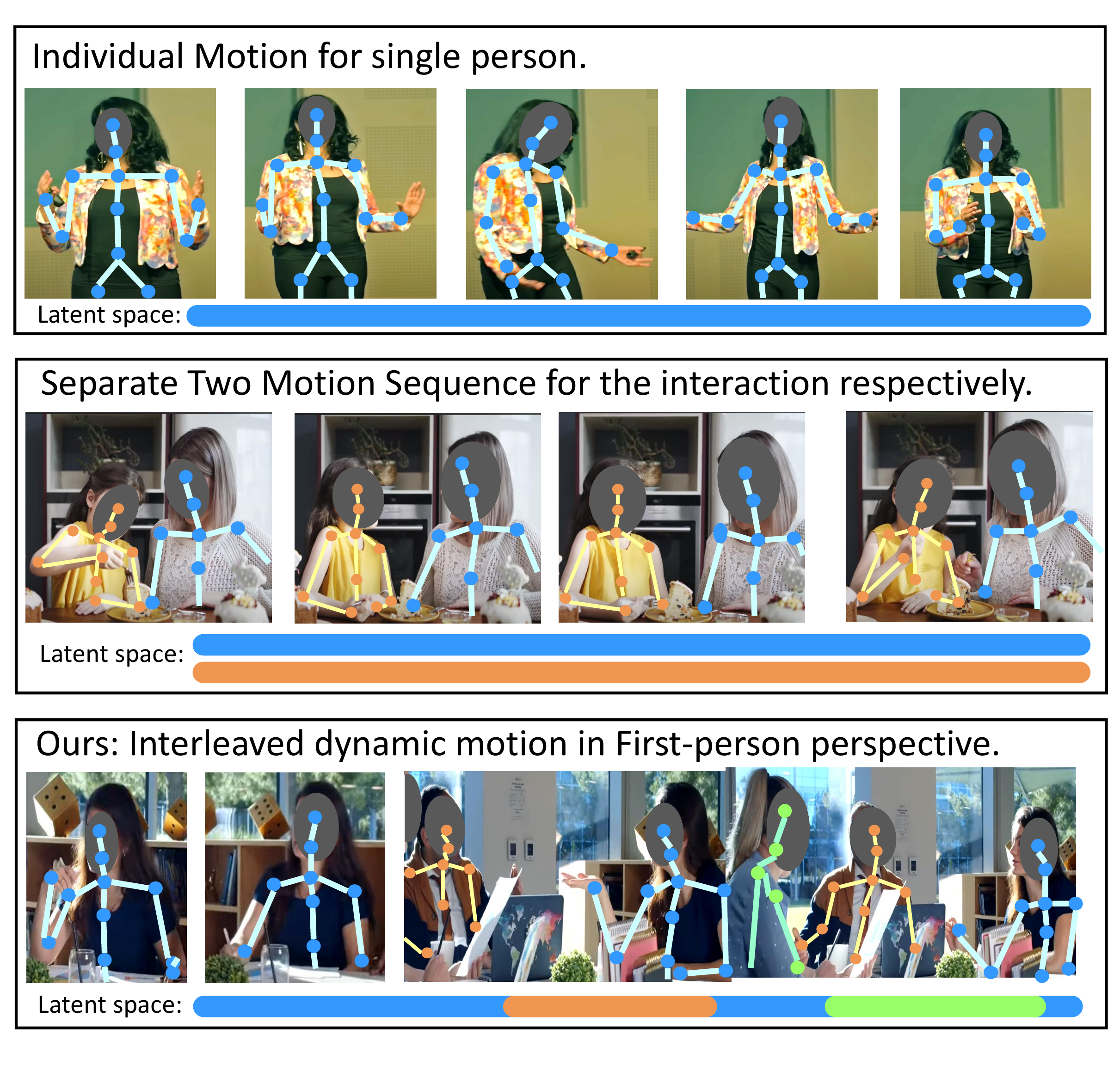}
    \caption{Unlike conventional approaches that process single-person and interaction motions separately, real human behavior exhibits a dynamic interleaving of solo and multi-person actions. Our InterSyn framework learns to generate these integrated motion sequences, closely emulating the fluid and natural dynamics observed in real-world interactions. For multi-human motion, InterSyn takes separate descriptions as inputs to better capture distinct roles and coordination.}
    \label{fig:motivation}
\end{figure}

Given the crucial role of motion information in applications such as visual perception~\cite{liang2019vrr,liang2022simple,liang2023icocap}, visual generation~\cite{zhang2024vast,chen2024hydra}, and 3D synthesis~\cite{wang2025structre,xi2025omnivdiff,huang2025mvtokenflow}, text-to-motion (T2M) synthesis~\cite{tevet2022human, jiang2024motiongpt, guo2024momask} has garnered increasing attention. Despite significant progress on benchmarks like HumanML3D~\cite{guo2022generating}, generating realistic and diverse human motion in the wild remains challenging. Current motion generators~\cite{tevet2023human, petrovich22temos, guo2022generating, liang2022seeg} often excel on specific datasets yet struggle to generalize to the varied and complex motions observed in real-world interactions. While some works~\cite{liang2024intergen, fan2024freemotion, zhong2024smoodi} have explored general motion generation for multiple characters, the generalization issue is still largely unresolved.

A promising direction for achieving motion generalization is to simulate real human learning. In everyday life, human motion arises from two intertwined components: semantic expression and social interaction. Motions for self-expression, such as walking and talking, are fundamental to individual development and well-being~\cite{inglehart2004individualism,sevdalis2012perceiving,mahfoudi2022emotion,izard1993four,isohatala2020cognitive,tomasello2005understanding}, while social motions like hugging or gesturing are primarily acquired through observation, imitation, and active engagement with others~\cite{bandura1977social,overskeid2018we,fryling2011understanding}. The dynamic nature of human learning, further supported by theories such as situated learning~\cite{sadler2009situated,herrington2000towards,utley2006effects,fors2013multisensory} and Cultural-Historical Activity Theory~\cite{igira2009cultural,engestrom2004new,nussbaumer2012overview}, as well as studies on peer group interactions~\cite{cohen2004social,lerner2018concepts,carter2005effects}, suggests that semantic and social motions mutually reinforce each other, providing a natural blueprint for more generalizable motion synthesis.

However, existing motion synthesis pipelines suffer from two critical limitations. First, they treat semantic and social motions as separate tasks—semantic actions such as walking or talking are typically learned in isolation (e.g., in HumanML3D~\cite{guo2022generating}), while social interactions are modeled independently (e.g., in InterGen~\cite{liang2024intergen}). As shown in Fig~\ref{fig:motivation}, this separation restricts the ability of models to generate the dynamic behaviors seen in natural settings. Second, many methods fail to capture the mutual cues of human motion, where individuals continuously adapt their movements in response to subtle signals from others. These dynamic adaptations are essential for producing vivid motion expressions.

To this end, we propose a novel framework, Interleaved Learning for Motion Synthesis (InterSyn), which learns to generate dynamic motion sequences by seamlessly integrating single-person and interaction motions. InterSyn comprises two key modules: the Interleaved Interaction Synthesis (INS) and the {Relative Coordination Refinement (REC). 
The INS module jointly models single-person and multi-person motions in a unified learning paradigm, enabling the learning of motion that akin to the real human being, who who change dynamically with solo motion and interaction motion seamlessly. Meanwhile, the REC module refines the synthesized interactions by leveraging the mutual dynamics and nuanced interdependencies among multiple characters, ensuring that the generated motions are both semantically coherent and socially synchronized.

Our extensive experiments demonstrate that InterSyn significantly outperforms existing methods in generating high-quality interaction motions. Quantitative evaluations on multiple datasets reveal substantial improvements in key metrics such as semantic coherence and synchronization between interacting individuals; for example, our method outperforms recent work~\cite{fan2024freemotion} by 40\% in terms of FID. Qualitative results further illustrate that our approach produces natural and nuanced motion sequences that capture the complex dynamics of real-world interactions. These findings confirm that by jointly modeling and refining solo and interactive dynamics, InterSyn effectively addresses the challenges of interaction motion synthesis, setting a new standard for robust and realistic motion generation.

Our contributions are threefold:
\begin{itemize}
    \item We introduce a novel framework, \textbf{InterSyn}, that jointly models single-person and interaction motions within a unified dynamic sequence, enabling the generation of realistic motions that naturally interleave individual and multi-person behaviors.
    \item We propose \textbf{Iterative Interaction Coordination (INC)} and \textbf{Relative Coordination Refinement (REC)}. INC facilitates dynamic motion generation by interleaving individual and interactive behaviors, while REC refines these interactions through relative coordination constraints that capture mutual dynamics among characters.
    \item Extensive experiments demonstrate that our approach significantly outperforms recent methods in generating diverse and realistic interactions, confirming the effectiveness of interleaved learning for natural motion synthesis.
\end{itemize}

\section{Related Work}
\label{sec:related}

Text-to-motion generation has gained significant attention for its applications in animation, virtual reality, and human-computer interaction. Research in this area has evolved from synthesizing realistic single-person motions to capturing complex multi-person interactions.

\textbf{Single-Person Motion Generation:} Diffusion-based approaches have emerged as powerful paradigms for text-to-motion synthesis. For example, the Motion Diffusion Model (MDM)~\cite{tevet2022human} adapts classifier-free diffusion by directly predicting the motion sample at each step, which facilitates the integration of geometric losses. Other works bypass diffusion by directly regressing pose sequences. {MoMask}~\cite{guo2024momask} introduces a hierarchical masked modeling framework that represents motion as multi-layer discrete tokens, where a Masked Transformer predicts masked tokens conditioned on text and a Residual Transformer refines these tokens, achieving notable improvements on benchmarks like HumanML3D and KIT-ML. In addition, methods such as {OmniControl}~\cite{xie2024omnicontrol} enable controllable motion synthesis via advanced conditioning techniques that allow users to specify attributes like speed and direction, while {TAAT}~\cite{wang2024taat} and emotion-enriched approaches leverage large language models (LLMs) to incorporate arbitrary texts and subjective motion details. 

\textbf{Multi-Person Motion Generation:} Moving beyond isolated actions, multi-person motion generation focuses on capturing the interactions and relationships among individuals. {SMooDi}~\cite{zhong2024smoodi} addresses stylized motion synthesis by combining content text with reference style sequences using a pre-trained text-to-motion model and a lightweight style adaptor. {FreeMotion}~\cite{fan2024freemotion} overcomes the limitations of fixed agent numbers by unifying single- and multi-person synthesis via a conditional motion distribution, enabling number-free generation with precise spatial control. Other approaches emphasize long-term, socially-aware interactions. {SATA}~\cite{mueller2024massively} employs a scene-aware social transformer that fuses motion and scene information within a diffusion framework, while {InterGen}~\cite{liang2024intergen} tailors diffusion-based generation to two-person interactions using cooperative transformer denoisers with mutual attention and a global relation representation.

\label{sec:method}

\section{Preliminary}

Diffusion models generate human motion through a probabilistic framework that iteratively denoises Gaussian noise into realistic sequences, extending their success in text-to-image synthesis \cite{tevet2023human}. Unlike traditional autoregressive methods, diffusion models synthesize full motion sequences holistically, addressing temporal coherence and physical plausibility.  

The process involves a forward noising stage that corrupts ground-truth motion $ x_0^* \in \mathbb{R}^{N \times D} $ (with $ N $ frames and $ D $-dimensional joint parameters) into Gaussian noise $ x_T $, and a learned reverse process that reconstructs $ x_0 $ from $ x_t $. At each diffusion step $ t $, the reverse transition is governed by:  
\begin{equation}  
P_{\theta}(x_{t-1} | x_t) = \mathcal{N}\left(\mu_t(\theta), \sigma_t I\right)
\end{equation}  
where $ \mu_t(\theta) $ combines the predicted clean motion $ x_0(\theta) $ and noisy input $ x_t $:  
\begin{equation}  
\mu_t(\theta) = \sqrt{\alpha_{t-1}} x_0(\theta) + \sqrt{1 - \alpha_{t-1} - \sigma_t^2} \cdot x_t
\end{equation}  
Here, $ \alpha_t \in (0,1) $ controls noise scheduling, with $ \beta_t = 1 - \alpha_t $, and $ \sigma_t^2 = \beta_t (1 - \alpha_{t-1}) / (1 - \alpha_t) $. The model directly predicts $ x_0 $ rather than incremental noise, optimizing via $ \mathcal{L} = \|x_0 - x_0^*\|_2^2 $.  

During inference, motion is synthesized by recursively denoising $ x_T \sim \mathcal{N}(0,I) $, leveraging cross-attention layers to align text embeddings with spatiotemporal motion features. This framework ensures physically plausible outputs while preserving semantic alignment with text prompts.

\begin{figure*}[t]
    \centering
    \includegraphics[width=\linewidth]{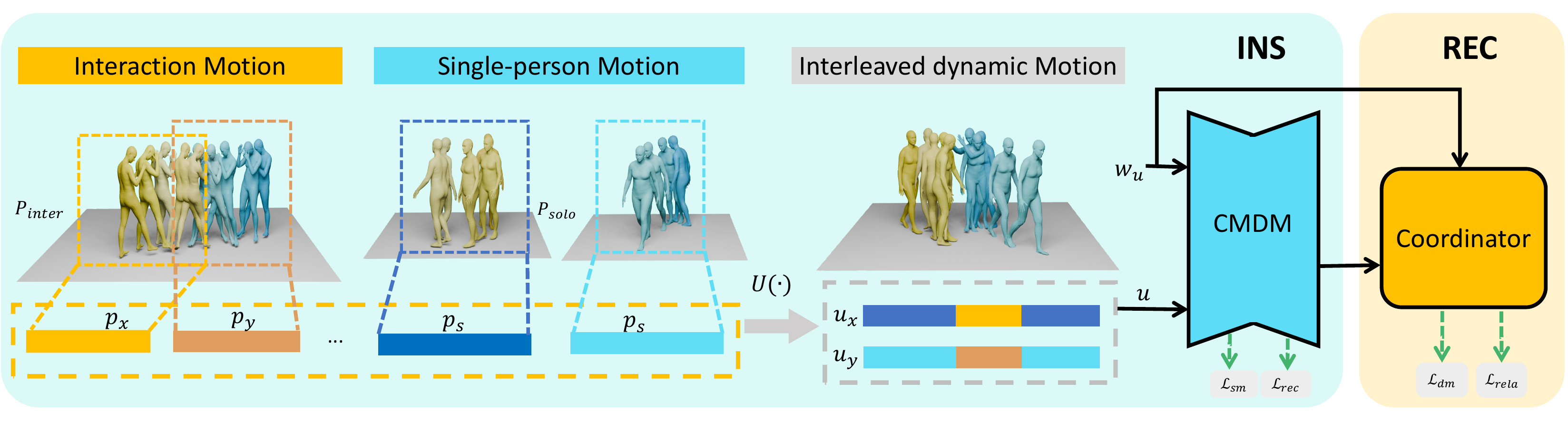}
    \caption{Overview of our interleaved learning framework (InterSyn) for dynamic motion synthesis. Our method consists of two stages: (1) Interleaved Interaction Synthesis (INS), which unifies single-person and interaction motions using a Conditional Motion Diffusion Model to generate a flexible, realistic motion schema, and (2) Relative Coordination Refinement (REC), which employs a Coordinator network to refine interactions based on mutual relationships among characters, yielding vivid and nuanced motion generation.}
    \label{fig:method}
\end{figure*}

\section{Interleaved Learning for Motion Synthesis}
\label{INS}
%In our research, the process of modeling conditional motion distribution is decomposed into two key components: the synthesis of single-person motion and the coordination of dual-person actions. To this end, we designed two main modules: Interleaved Interaction Synthesis (INS) and Relative Coordination Refinement (REC). The INS module is primarily responsible for generating single-agent motion. It can synthesize reasonable single-person motion sequences based on the input conditional information. On the other hand, the REC module handles the coordination of multiple agents' motions. It primarily addresses the interaction problem between two or more agents under given external control conditions, ensuring that the motion sequences can interact with each other in a reasonable manner to avoid collisions or conflicts. By precisely controlling the relative positions and motion synchronization between characters, the motion coordination module effectively enables natural interactions between multiple agents.
\subsection{Interleaved Interaction Synthesis}

Human motion comprises both individual movements (for self-expression) and multi-person interactions (for social engagement). In natural settings, people fluidly transition between these modes, adapting their actions to the context. To capture this behavior, we propose the Interleaved Interaction Synthesis (INS) framework, which jointly models single-person and interaction motions in a unified paradigm. Unlike existing methods that treat these motion types independently, INS learns them together, enabling seamless transitions between solo and interactive actions and closely mimicking the fluidity of real human movement.

We begin by processing and unifying all single-person and interaction motions from a first-person perspective. Specifically, we initialize a motion \emph{bucket} $u$ to represent the interleaved sequence, where
\begin{equation}
 u = ( u_x,u_y) \in \mathbb{R}^{2 \times T \times K \times C}
\end{equation}
with $T$ denoting the number of time steps, $K$ the number of keypoints per motion, and $C$ the dimensionality of each keypoint. To construct $u$ that adaptively interleaves single-person and multi-person motions, we first randomly sample a single-person motion $p_s$ from a diverse dataset~\cite{tevet2022human}. 

Then, for multi-person interactions, let
\[
P_{i} = \{p_1, p_2, \ldots, p_N\}
\]
denote the set of motions from $N$ individuals. We randomly select a pair of interaction motions $(p_x, p_y)$ to interleave with $p_s$. These motions are combined into a unified sequence using a function $U(\cdot)$:
\begin{equation}
    u = U(p_x, p_y, p_s, t_i, t_s)
\end{equation}
where $t_i$ and $t_s$ are the start time indices for the interaction motion $(p_x,p_y)$ and the solo motion $p_s$, respectively. The sequence $u_x$ can begin with either $p_x$ or $p_s$ (ensuring $t_i + t_s > 0$ and $t_i \cdot t_s = 0$) so that exactly one segment starts at the first time step. The $U(\cdot)$ manages motion alignment, smooth transitions, and orientation adjustments between segments. Based on this, "interleave" means blending multi-person interaction into single-person motions using a function that alternates between solo and interactive actions. 

Due to the mismatch of skeletons in the fusion of two datasets, we need to apply rotation and translation to the skeleton $ sk_2 $ relative to $ sk_1 $ in order to make their expression of the motion consistent. In forward kinematics (FK), the goal is to determine the global position of each joint in a skeletal structure based on its local transformations, which consist of rotations and translations. Let the global position of the $ k $-th joint in skeleton $ sk_1 $ be denoted as $ j_k^{sk_1} $, and in skeleton $ sk_2 $ as $ j_k^{sk_2} $. Transforming the motion from one skeleton to another requires adjusting both the bone lengths and rotations of the joints.

The process starts with the root joint, which serves as the starting point for all subsequent transformations. We assume that the global position of the root joint is the same in both skeletons. For each joint $ k $, the relative position from its parent joint in both skeletons is adjusted by scaling the bone length and applying the rotational difference. In forward kinematics, starting from the root node and extending to the child nodes, the transformation from $ sk_2 $ to $ sk_1 $ can be expressed as:

\begin{equation}
    j_k^{sk_1} = R_i \cdot j_{k-1}^{sk_2} + T_i (j_{k}^{sk_2} - j_{k-1}^{sk_2})
\end{equation}
where $ R_i $ is the rotation matrix for the $ i $-th joint relative to the parent joint $ j_{i-1} $ in $ sk_2 $ and $ sk_1 $, and $ T_i $ is the translation matrix corresponding to the bone length difference between the $ i $-th joint in $ sk_2 $ and $ sk_1 $ relative to the parent joint $ j_{i-1} $. More details about the $U(\cdot)$ can be found in the appendix.

Moreover, text descriptions for $p_s$, $p_x$, and $p_y$ are denoted as $w_s$, $w_x$, and $w_y$. We concatenate $w_s$ and $w_x$ into a combined description $w_u$ for $u$.

To synthesize the dynamic motion $u$, we employ a Conditional Motion Diffusion Model (CMDM), denoted as $M_s$. As shown in Fig.~\ref{fig:method}, $M_s$ is a Transformer-based diffusion network~\cite{tevet2022human} that incorporates multiple denoising blocks. The model takes the interaction time steps $t_i$ and $t_s$ as conditioning inputs and includes time embedding layers to encode $t_i$ and $t_s$, as well as a text embedding layer to encode $w_u$. During training, at time step $t$, we add noise to the motion $u$ to obtain a noisy motion $u^t$, and concatenate the resulting embeddings, which are then fed into the diffusion model:
\begin{equation}
    \hat{u} =(\hat{u_x},\hat{u_y})= M_s(u, w_u, t_i, t_s)
\end{equation}
By conditioning on these temporal signals, the diffusion model learns to generate a continuous motion sequence that smoothly interleaves individual and interactive actions.

We optimize INS using the following loss function:
\begin{equation}
    \mathcal{L}_{I} = \lambda_1 \mathcal{L}_{\text{rec}} + \lambda_2 \mathcal{L}_{\text{smooth}}
\end{equation}
where $\mathcal{L}_{\text{rec}}$ is the reconstruction loss of the CMDM for $p_u$, and $\mathcal{L}_{\text{smooth}}$ enforces smooth transitions at the boundaries (within a window of $\pm5$ frames) during motion fusion. The hyperparameters $\lambda_i$ balance these loss terms.
\begin{equation}
   \mathcal{L}(u)_{\text{smooth}} = \sum_{t = t_i - 5}^{t_i + 5} \left| \frac{d}{dt} \left( u(t) \right) \right| 
\end{equation}

While our framework supports multiple interleaved segments (by introducing additional $t_i$ and $t_s$ ), we observe that more than two segments per sequence degrade performance. This is because increasing the number of segments reduces the duration allocated to each motion, potentially truncating key motion phases and hindering semantic capture.

\subsection{Relative Coordination Refinement}

Interaction motions require both individual semantic coherence (each person’s motion is contextually meaningful) and synchronized social coordination (movements are responsive and harmonized). To address this, we propose the Relative Coordination Refinement (REC) module, which refines multi-person interactions by considering spatial relationships and mutual influence among characters.

In REC, a coordinator network $M_c$ (see Fig.~\ref{fig:method}) refines the interaction predictions. We use the Transformer network architecture to build the coordinator module. The Transformer is primarily used to model the spatial and temporal relationships between multiple motion sequences. During training phase, for example, in a two-person interaction as in~\cite{liang2024intergen}, the first person's predicted motion can be refined using the second's motion, and vice versa. Let $\hat{u_y}$ be the reference motion. The refinement process for $\hat{u_x}$ is as follows:
\begin{align}
    % \hat{u_x}&= M_s(u, w_u, t_i, t_s), \\
    \phi_x &= M_c\bigl(\hat{u}_x,\, \hat{u}_y,\, w_u\bigr)
\end{align}
where $\hat{u_x}$ is the motion reconstructed by $M_s$ via the diffusion process and is further refined by $M_c$ to yield $\phi_x$.

Similarly, we process the reference interaction motion $u_y$ by passing it through the diffusion model $M_s$ without applying any temporal offset. We obtain:
\begin{align}
    \phi_y &= M_c\bigl(\hat{u}_y,\, \phi_x,\, w_u\bigr)
\end{align}

We model the problem of handling social coordination as the conditional distribution $P(\hat{u_x} | \hat{u_y})$  , where the model needs to learn how to process the action sequence  $\hat{u_x}$  based on the spatial information of  $\hat{u_y}$. We simulate the $ M_c$  module as a mechanism that performs translation and rotation on the spatial positions of $\hat{u_x}$'s actions, as well as a rationalization of its trajectory, ensuring that $\hat{u_y}$ can interact appropriately with  $\hat{u_x}$. 

Therefore, when $\phi_x$  represents a reasonable interaction action relative to $\hat{u_y}$, we fine-tune $\hat{u_y}$ conditionally to obtain $\phi_y$. Based on the functional definition of the $ M_c$ module, $\phi_y$ should be infinitesimally close to $\hat{u_y}$. Based on this, we define the relative coordination loss as:
\begin{equation}
    \mathcal{L}_{\text{rela}}(\phi_y, \hat{u_y}) =  \| \phi_y - \hat{u_y} \|_2
\end{equation}

The REC loss is formulated as:
\begin{equation}
    \mathcal{L}_{R} = \lambda_3\mathcal{L}_{\text{rela}}(\phi_y, \hat{u_y}) + \lambda_4\, \mathcal{L}_{\text{dm}}(\phi_x, \hat{u_y})
\end{equation}
where $\mathcal{L}_{\text{dm}}$ is the masked joint distance map loss from~\cite{liang2024intergen}.

Finally, the overall loss function is:
\begin{equation}
    \mathcal{L} = \mathcal{L}_I + \mathcal{L}_R
\end{equation}

\section{Experiment}
\label{sec:experiment}

\textbf{Datasets.} We conducted experiments on the \textbf{HumanML3D} \cite{guo2022generating} single-human dataset and the \textbf{InterHuman} \cite{liang2024intergen} dual-human dataset. The HumanML3D dataset re-annotates the AMASS dataset \cite{mahmood2019amass} and the HumanAct12 dataset \cite{guo2020action2motion}, offering a total of 44,970 annotations across 14,616 motion sequences. The InterHuman dataset is the first dual-human motion dataset with text annotations, containing 6,022 different actions across various human activity types, each action accompanied by 16,756 unique textual descriptions, using 5,656 different words.

\textbf{Evaluation Metrics.} We follow the evaluation protocol outlined by \cite{guo2022generating}. To evaluate the naturalness of the generated motions, we utilize the Frechet Inception Distance (FID). R-Precision is employed to assess the alignment between the generated motions and the text prompt, while Diversity quantifies the variation within the generated motions. Multimodality (MModality) captures the diversity present within a single text prompt. Finally, Multi-modal distance (MM Dist) measures the distance between the motion and text feature representations.

\subsection{Implementation Details}
During training, we adopt an alternating training strategy using interleaved data and HumanML3D data. Specifically, within a single learning process, the model computes the reconstruction loss $\mathcal{L}_{\text{rec}}$ not only for the single-person data but also for the data obtained by fusing the single-person data with InterHuman. This approach ensures that the CMDM model learns more efficient denoising capabilities from the single-person data training, as the interleaved data $u$ inherently contains noise from the data processing. For the REC module, we use transformer. 

The diffusion timesteps are set to 1,000, and the DDIM \cite{song2020denoising} sampling strategy is used during the inference stage. In the first stage, the INS module is trained, and the overall loss function is \( L_{\text{ins}} \). In the second stage, the parameters of the INS module are frozen, and the INC module is trained, with the total loss being \( L \). The hyperparameters are set as \( \lambda_1 = 1 \),  \( \lambda_2 = 0.1 \),  \( \lambda_3 = 1\), and \( \lambda_4 = 0.5 \). Both training stages use a learning rate of \( 1e-5 \), with 2,000 training epochs for each stage. We use the frozen CLIP-ViT-L-14 model as the text encoder. Our model trains in 31 hours on a single H100 GPU with a batch size of 256 and 44 GB of memory. For the data ratio issue between the two datasets, please refer to the appendix for details.

For the function $U(\cdot)$, we adopt the data representation method from HumanML3D, where each frame of motion is represented as a 263-dimensional feature vector. All single-person and dual-person data are uniformly normalized for initial positions, orientations, and skeletal frameworks, and the data are processed into a recursive form.

To make a fair comparison, we benchmarked FreeMotion~\cite{fan2024freemotion}, which is also dedicated to tackling the issues of number-free motion generation and human-to-human interaction. FreeMotion trains its number-free motion generation model by using InterHuman dataset. As shown in Table \ref{tab1}, we adopted the same training data format as FreeMotion and conducted tests on the relevant metrics. At the same time, since the FreeMotion framework cannot train on single-human datasets, we used $U(\cdot)$ function at the input stage, consistent with InterSyn, for processing and training. 
\subsection{Quantitative Results}

\begin{figure*}[tp]
    \centering
    \includegraphics[width=\linewidth]{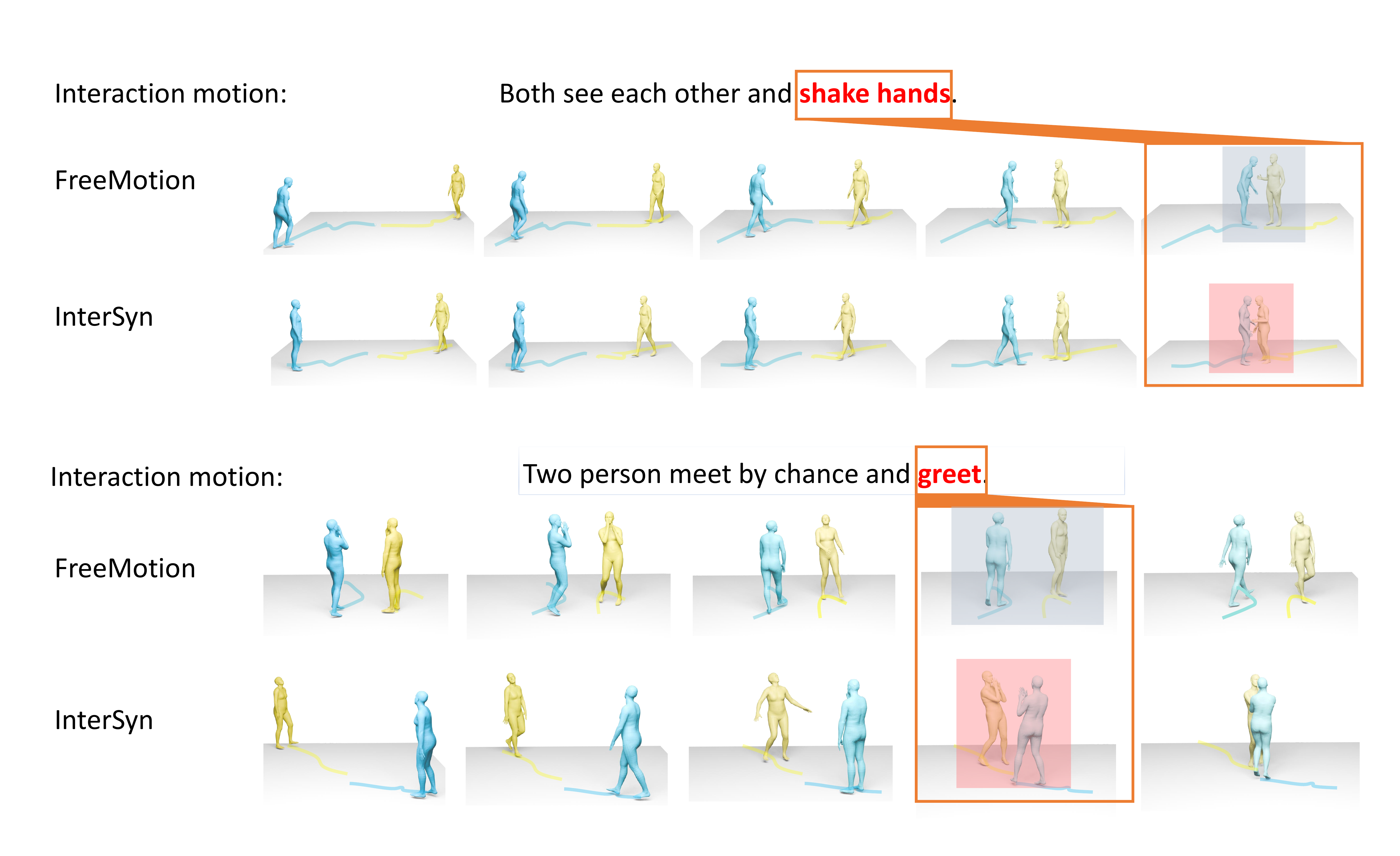}
    \caption{Example of generated interaction motions. By learning from interleaved motions that closely mirror real human dynamics, our method produces more vivid and natural interactions. Significant interaction regions are highlighted in blue for \cite{fan2024freemotion} and in red for our approach.}
    \label{vis1}
\end{figure*}

\begin{figure*}[tp]
    \centering
    \includegraphics[width=\linewidth]{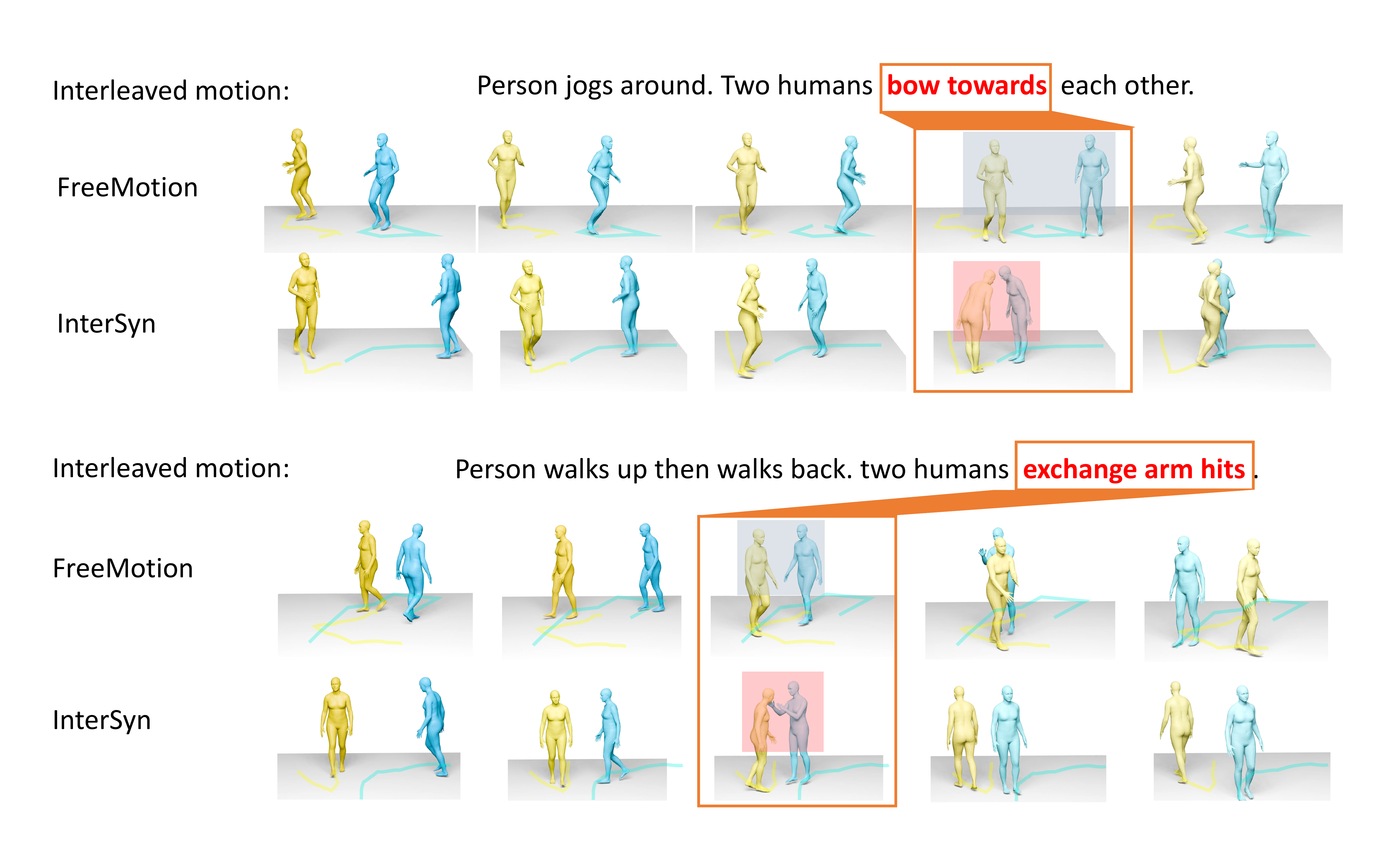}
    \caption{Example of generated interleaved motions. Our method produces realistic and nuanced sequences that closely mimic real-world scenarios by seamlessly integrating solo actions and interactions. The visualized motions highlight improved fidelity and smooth transitions compared to prior work. Significant interaction regions are marked in blue for \cite{fan2024freemotion} and in red for our approach.}
    \label{vis}
\end{figure*}

\begin{table*}[t]
\centering
% \resizebox{\columnwidth}{!}{%

\begin{tabular}{lccccccc}
\toprule
\multirow{2}{*}{Method} & \multicolumn{3}{c}{R Precision $\uparrow$} & \multirow{2}{*}{FID $\downarrow$}& \multirow{2}{*}{MM Dist $\downarrow$} & \multirow{2}{*}{Diversity $\rightarrow$} & \multirow{2}{*}{ MModality $\uparrow$} \\
                        & \centering Top 1 & \centering Top 2 & \centering Top 3 & & & & \\

\midrule
Real  & $0.452^{\pm .008}$ & $0.610^{\pm .009}$ & $0.710^{\pm .008}$  & $0.273^{\pm .007}$ & $3.755^{\pm .008}$ & $7.748^{\pm .064}$ & - \\
\midrule
TEMOS~\cite{petrovich2022temos}  & $0.224^{\pm .010}$ & $0.316^{\pm .013}$ & $0.450^{\pm .018}$  & $17.375^{\pm .043}$ & $5.342^{\pm .015}$ & $6.939^{\pm .071}$ & $0.535^{\pm .014}$ \\
T2M~\cite{guo2022generating}  & $0.238^{\pm .012}$ & $0.325^{\pm .010}$ & $0.464^{\pm .014}$  & $13.769^{\pm .072}$ & $4.731^{\pm .013}$ & $7.046^{\pm .022}$ & $1.387^{\pm .076}$ \\
MDM~\cite{tevet2023human} & $0.153^{\pm .012}$ & $0.260^{\pm .009}$ & $0.339^{\pm .012}$ & $9.167^{\pm .056}$ & $6.125^{\pm .018}$ & $7.602^{\pm .045}$ & $\mathbf{2.355^{\pm .080}}$ \\
ComMDM~\cite{shafir2023human} & $0.223^{\pm .009}$ & $0.334^{\pm .008}$ & $0.466^{\pm .010}$ & $7.069^{\pm .054}$ & $5.212^{\pm .021}$ & $7.244^{\pm .038}$ & $1.822^{\pm .052}$ \\
InterGen~\cite{liang2024intergen} & $0.264^{\pm .006}$ & $0.392^{\pm .005}$ & $0.472^{\pm .005}$ & $13.404^{\pm .200}$ & $3.882^{\pm .001}$ & $7.77^{\pm .030}$ & $1.451^{\pm .034}$ \\
FreeMotion~\cite{fan2024freemotion} & $0.326^{\pm .003}$ & $0.462^{\pm .006}$ & $0.544^{\pm .006}$ & $6.740^{\pm .130}$ & {$\mathbf{3.848^{\pm .002}}$} & $7.828^{\pm .130}$ & $1.226^{\pm .046}$ \\
\midrule
Ours & $\mathbf{0.335^{\pm .008}}$ & $\mathbf{0.479^{\pm .006}}$ & $\mathbf{0.584^{\pm .008}}$ & {$\mathbf{6.332^{\pm .210}}$} & $3.856^{\pm .005}$ & $\mathbf{7.763^{\pm .052}}$ & {$1.601^{\pm .047}$}\\
\bottomrule
\end{tabular}
% }
\caption{Quantitative comparisons on the InterHuman test set. Our method significantly outperforms competing approaches across various metrics.}

\vspace{-0.05in}
\label{tab1}
\end{table*}

\textbf{Comparisons with state-of-the-art models: } We first compare the interaction performances in InterHuman test set. As in Table~\ref{tab1}, our method achieves state-of-the-art performance across all metrics, demonstrating superior text-to-motion synthesis capabilities. With R Precision scores of 0.335, 0.479, and 0.584 for top-1, top-2, and top-3 respectively, and a low FID of 6.332, InterSyn excels in translating textual semantics into physically plausible interactions. The INS module effectively resolves ambiguities in action descriptions by jointly modeling individual and interactive motions, ensuring precise alignment between text prompts and spatiotemporal dynamics. Notably, our approach outperforms InterGen by achieving a 25.3\% improvement in Top-1 R Precision and a 50.6\% reduction in FID, highlighting its superior capability in capturing nuanced coordination. Furthermore, our framework achieves near-real diversity (7.763 vs. 7.748) and optimal MModality (1.601), confirming its ability to generate varied yet natural interactions. The REC module enforces interaction aware constraints while preserving motion variety, and compared to single-person methods such as MDM, InterSyn reduces MM Dist by 37.0\%, underscoring its effectiveness in synthesizing geometrically coherent multi-actor motions. Overall, these results validate that interleaved learning of solo and interactive dynamics—coupled with explicit coordination refinement—is critical for generating socially synchronized motions that accurately reflect textual intent while maintaining human-like naturalness.

\textbf{Evaluation in Dynamic Environments: } Human motion synthesis must address the inherent complexity of real-world scenarios where solo actions and social interactions coexist and dynamically interleave. To rigorously evaluate this capability, we construct an interleaved motion benchmark unifying HumanML3D and InterHuman test sets, reflecting natural transitions between individual and interactive behaviors. To avoid alignment issues from fused datasets, we use a hybrid strategy: single-person and multi-person evaluators are applied to their respective segments, and for mixed test sequences, we compute a frame-wise weighted average of the matching scores. Temporal masking is applied to boundary frames to prevent bias from transitions. As shown in Table~\ref{tab:2}, our method achieves a 6.4\% improvement in Top 1 R Precision and 42.1\% reduction in FID over FreeMotion. These gains stem from two points:
(1). InterSyn jointly encodes solo and interactive motions within a unified latent space, enabling seamless transitions between action types. This eliminates the modality bias observed in FreeMotion, which processes individual and group motions through separate pathways, leading to inconsistent semantics during dynamic interleaving (e.g., abrupt posture breaks when switching from `walking alone' to `shaking hands').
(2). Our method reduces MM Dist by 5.1\% through adaptive kinematic constraints that preserve spatial relationships during motion transitions. For instance, it maintains appropriate interpersonal distances when synthesizing sequences like `sit down → converse → stand up', where FreeMotion often introduces implausible body collisions.

The method also achieves a 13.3\% higher MModality, demonstrating its capacity to generate diverse and text-consistent motions. This is exemplified in scenarios like `passing an object while walking', where our framework produces varied arm trajectories and walking speeds while ensuring object transfer completion—a nuance that FreeMotion fails to capture due to its decoupled motion modeling. Notably, our approach attains this without sacrificing diversity, proving that joint modeling enhances rather than restricts motion variety. In all, the results validate that interleaved learning of solo and interactive dynamics is essential for synthesizing motions in real-world environments, where actions fluidly blend individual intent and social coordination.

\begin{table*}[t]
\centering
% \resizebox{\columnwidth}{!}{%

\begin{tabular}{lccccccc}
\toprule
\multirow{2}{*}{Method} & \multicolumn{3}{c}{R Precision $\uparrow$} & \multirow{2}{*}{FID $\downarrow$}& \multirow{2}{*}{MM Dist $\downarrow$} & \multirow{2}{*}{Diversity $\rightarrow$} & \multirow{2}{*}{ MModality $\uparrow$} \\
                        & \centering Top 1 & \centering Top 2 & \centering Top 3 & & & & \\

\midrule
FreeMotion~\cite{fan2024freemotion} & $0.280^{\pm .011}$ & $0.381^{\pm .008}$ & $0.508^{\pm .010}$ & $0.721^{\pm .016}$ & $3.905^{\pm .002}$ & $9.244^{\pm .091}$ & $1.397^{\pm .066}$ \\
\midrule
ours & $\mathbf{0.298^{\pm .010}}$ & $\mathbf{0.410^{\pm .007}}$ & $\mathbf{0.508^{\pm .008}}$ & $\mathbf{0.417^{\pm .012}}$ & $\mathbf{3.707^{\pm .003}}$ & $\mathbf{9.322^{\pm .030}}$ & $\mathbf{1.583^{\pm .029}}$\\
\bottomrule
\end{tabular}

\caption{Quantitative comparisons on our interleaved motion dataset, constructed by unifying the HumanML3D and InterHuman test sets to better reflect real-world interactions. Our method demonstrates significant improvements across all evaluated metrics.}

\vspace{-0.05in}
\label{tab:2}
\end{table*}

\begin{table*}[t]
\centering
% \resizebox{\columnwidth}{!}{%

\begin{tabular}{lccccccc}
\toprule
\multirow{2}{*}{Method} & \multicolumn{3}{c}{R Precision $\uparrow$} & \multirow{2}{*}{FID $\downarrow$}& \multirow{2}{*}{MM Dist $\downarrow$} & \multirow{2}{*}{Diversity $\rightarrow$} & \multirow{2}{*}{ MModality $\uparrow$} \\
                        & \centering Top 1 & \centering Top 2 & \centering Top 3 & & & & \\

\midrule
$s-i-s$ & $0.298^{\pm .010}$ & $0.410^{\pm .007}$ & $0.508^{\pm .008}$ & $0.417^{\pm .012}$ & $3.707^{\pm .003}$ & $9.322^{\pm .030}$ & $1.583^{\pm .029}$\\

$s-i-s-i$ & $0.242^{\pm .008}$ & $0.301^{\pm .010}$ & $0.477^{\pm .004}$ & $0.469^{\pm .024}$ & $3.958^{\pm .011}$ & $9.870^{\pm .056}$ & $1.637^{\pm .071}$\\

$i-s-i-s$ & $0.238^{\pm .011}$ & $0.319^{\pm .005}$ & $0.453^{\pm .003}$ & $0.467^{\pm .041}$ & $3.880^{\pm .020}$ & $9.634^{\pm .038}$ & $1.600^{\pm .056}$\\

$s-i-s-i-s$ & $0.115^{\pm .006}$ & $0.189^{\pm .003}$ & $0.267^{\pm .003}$ & $0.638^{\pm .055}$ & $4.436^{\pm .010}$ & $11.493^{\pm .118}$ & $1.837^{\pm .109}$ \\

$i-s-i-s-i$ & $0.116^{\pm .010}$ & $0.204^{\pm .005}$ & $0.318^{\pm .007}$ & $0.661^{\pm .024}$ & $4.519^{\pm .015}$ & $12.945^{\pm .064}$ & $1.941^{\pm .050}$\\

\midrule
$t_i = 0, t_s = 60$ & $0.301^{\pm .003}$ & $0.411^{\pm .009}$ & $0.519^{\pm .006}$ & $0.422^{\pm .009}$ & $3.695^{\pm .007}$ & $9.036^{\pm .082}$ & $1.416^{\pm .052}$\\

$t_i = 0, random \, t_s$ & $0.283^{\pm .012}$ & $0.398^{\pm .007}$ & $0.503^{\pm .004}$ & $0.418^{\pm .020}$ & $3.710^{\pm .004}$ & $9.271^{\pm .049}$ & $1.479^{\pm .037}$\\

$t_s = 0, t_i = 60$ & $0.310^{\pm .005}$ & $0.426^{\pm .005}$ & $0.521^{\pm .011}$ & $0.423^{\pm .012}$ & $3.702^{\pm .005}$ & $9.102^{\pm .074}$ & $1.399^{\pm .058}$ \\

$t_s = 0, random \,t_i$  & $0.298^{\pm .010}$ & $0.410^{\pm .007}$ & $0.508^{\pm .008}$ & $0.417^{\pm .012}$ & $3.707^{\pm .003}$ & $9.322^{\pm .030}$ & $1.583^{\pm .029}$\\

\midrule
w/o coordinator & $0.103^{\pm .009}$ & $0.178^{\pm .008}$ & $0.335^{\pm .021}$ & $0.847^{\pm .104}$ & $5.842^{\pm .018}$ & $11.270^{\pm .142}$ & $1.837^{\pm .044}$ \\
w/o $\mathcal{L}_{\text{rela}}$ & $0.283^{\pm .023}$ & $0.393^{\pm .012}$ & $0.511^{\pm .015}$ & $0.537^{\pm .021}$ & $3.838^{\pm .007}$ & $9.370^{\pm .072}$ & $1.543^{\pm .059}$ \\
w/o $\mathcal{L}_{\text{smooth}}$ & $0.295^{\pm .020}$ & $0.399^{\pm .008}$ & $0.509^{\pm .013}$ & $0.431^{\pm .031}$ & $3.712^{\pm .013}$ & $9.312^{\pm .053}$ & $1.544^{\pm .037}$ \\

\bottomrule
\end{tabular}
\caption{Ablation study evaluating the impact of data organization for interleaved motion, INC timestep settings, and loss function designs.}

\vspace{-0.05in}
\label{tab:3}
\end{table*}

\textbf{Ablation on Data orgnization of INC: }
As shown in Table~\ref{tab:3}, after setting the fixed frame number to 196 for fair comparison, we conducted experiments using different data ratios and sequencing schemes for single- and dual-human interaction data (e.g., s-i-s-i'' represents a sequence where the data alternates between single-person and interactive data segments). From the table, we can observe that incorporating more complex and frequently switching data patterns within a limited time window leads to a decline in the model's learning capacity, ultimately resulting in poorer performance. In contrast, the simpler method (s-i-s'') not only yields better motion expressiveness but also proves to be more robust and scalable for long-duration interaction modeling tasks.

\textbf{Ablation for timestep settings in INC: }
Based on the ablation study results in Table \ref{tab:3}, we can conclude that the configuration of $t_s = 0, random , t_i$ performs the best across multiple metrics. It achieves a better balance in Top 1 precision, Diversity, and MModality compared to other settings like $t_i = 0, t_s = 60$ or $t_s = 0, t_i = 60$. Specifically, in terms of Diversity and MModality, $t_s = 0, random , t_i$ shows higher values, indicating that it generates more diverse and multi-modal action sequences. Therefore, $t_s = 0, random , t_i$ is considered the optimal configuration, as it adapts better to different time settings and complex interaction tasks, delivering improved model performance and generation diversity.

% \noindent 
% According to Table \ref{tab:5}, the results show that each component (coordinator, $\mathcal{L}{\text{rela}}$, and $\mathcal{L}{\text{smooth}}$) contributes to improving performance, with the full model achieving the best overall results in terms of R Precision, FID, MM Dist, Diversity, and Modality.

\textbf{Ablations on loss functions: }
We systematically evaluate the loss function of InterSyn as in Table~\ref{tab:3}. 
In REC, removing the coordination module (w/o coordinator) severely degrades performance, with MM Dist increasing by 57.6\% and FID by 103.1\%. This demonstrates the module's critical role in improving the interaction motions. Moreover, disabling $\mathcal{L}_{\text{rela}}$ reduces Top 1 R Precision by 5.0\%, revealing its specific function in aligning mutual dynamics of multi-character interactions (e.g., maintaining proper arm positions during "hugging" actions). While ablation of $\mathcal{L}_{\text{smooth}}$ shows minimal metric impact (+3.4\% FID), qualitative analysis reveals jittery transitions in motion sequences (e.g., abrupt footplant shifts during gait changes), confirming its role in ensuring natural motion flow. 

\subsection{Qualitative Evaluation}

\textbf{Visualization of Interaction Motions:}
Our proposed method demonstrates significant improvements over the baseline FreeMotion approach, as evidenced by the visual results. The visualizations clearly highlight the enhanced performance in generating more natural interactions and better text matching. As shown in Fig.~ \ref{vis1} and Fig.~ \ref{vis}, the free motion model still exhibits significant deficiencies in generating interactive motions for two-person actions, and its framework does not generalize well to dual-agent motion interaction tasks. In contrast, InterSyn generates motions with higher text matching accuracy, whether using single-person or two-person text inputs. Furthermore, the results demonstrate that InterSyn achieves controllability over the generated outcomes through \( t_i \) and \( t_s \), allowing for the customization of motion sequence details and the generation of corresponding actions.

\textbf{Visualization on Dynamic motions:}
Using the design structure of the coordinator, the model is able to extend the generation of two-person motions to multiple agents during the generation phase. The final results and discussion can be found in the appendix.

\section{Conclusion}
\label{sec:conclusion}

In this paper, we introduced {Interleaved Learning for Motion Synthesis (InterSyn)}, a novel framework that unifies the generation of dynamic interaction motions by seamlessly integrating single-person and multi-person dynamics. Our approach leverages the {Interleaved Interaction Synthesis (INS)} module to jointly model motion, capturing the fluid transitions between solo and interactive behaviors, and the {Relative Coordination Refinement (REC)} module to fine-tune interdependent interactions, ensuring semantic coherence and social synchronization. Extensive experiments validate that InterSyn produces more realistic and nuanced interaction motions than state-of-the-art methods. 

\noindent\textbf{Acknowledgement.} This work was partly supported by the Shenzhen Key Laboratory of Next Generation Interactive Media Innovative Technology (Project No. ZDSYS20210623092001004).

{\small
\bibliographystyle{ieeenat_fullname}
\bibliography{11_references}
}

%\clearpage 
\appendix 
\section{Appendix Section}
\label{sec:appendix_section}
In the following sections, we present detailed insights into the design and performance of our method for multi-person motion generation and coordination. The Design Details of the $U(\cdot)$ Function section elaborates on the key aspects of motion alignment, smooth transitions, and orientation adjustments, which are crucial for generating coherent and realistic motion sequence. More additional experiments results are displayed in the Additional Experiments section. In the Multi-Agent Motion Generation Results and Analysis section, we demonstrate how the dual-person interaction alignment module can be extended to multiple agents, illustrating the process and limitations of generating interactive motion sequences for more complex multi-agent scenarios. Finally, the Additional Visual Results section provides a series of visualizations that highlight the model's effectiveness in synthesizing and coordinating diverse motion sequences across different scenarios, offering further insight into its ability to handle complex interactions and maintain synchronization between multiple agents.
\subsection{Design Details of the $ U(\cdot) $ Function}
The function $U(\cdot)$ manages motion alignment, smooth transitions, and orientation adjustments between segments, resulting in a coherent interleaved sequence. 

\textbf{Motion Alignment.} The transformation from $ sk_2 $ to $ sk_1 $ can be expressed as:
\begin{equation}
    j_k^{sk_1} = R_k \cdot j_{k_1}^{sk_2} + T_k j_{k}^{sk_2}+\left(1 -T_k\right) j_{k-1}^{sk_2}
\end{equation}
where $ R_k $ is the rotation matrix for the $ k $-th joint relative to the parent joint $ j_{k-1} $ in $ sk_2 $ and $ sk_1 $, and $ T_k $ is the translation matrix corresponding to the bone length difference between the $ k $-th joint in $ sk_2 $ and $ sk_1 $ relative to the parent joint $ j_{k-1} $. 

Given the quaternions \( q_k^{sk_1} \) and \( q_k^{sk_2} \), which represent the rotations of the \( k \)-th joint relative to its parent joint \( j_{k-1} \) in skeletons \( sk_1 \) and \( sk_2 \), respectively, the relative rotation quaternion \( q_i \) between \( sk_1 \) and \( sk_2 \) is computed as:

\begin{equation}
q_k = q_k^{sk_1} \cdot (q_k^{sk_2})^{-1}
\end{equation}
where \( (q_i^{sk_2})^{-1} \) is the inverse of \( q_i^{sk_2} \), equivalent to the conjugate of \( q_i^{sk_2} \). The rotation matrix \( R_k \) can then be obtained from the quaternion \( q_i \) using the quaternion-to-rotation matrix conversion:
\begin{equation}
   R_k = \text{QuatToRot}(q_i)
\end{equation}
where \( q_i = (w_i, v_i) \) with \( w_i \) being the scalar part and \( v_i = (v_{i1}, v_{i2}, v_{i3}) \) being the vector part of the quaternion.

To align the bone lengths between two skeletons at the $k$-th joint relative to the parent joint, we define $T_k$ as the ratio of the bone length in $sk_1$ to the bone length in $sk_2$. This ratio accounts for the difference in bone lengths between the corresponding joints in the two skeletons.

The formula for $T_k$ can be expressed as follows:

\begin{equation}
    T_k = \frac{L_k^{sk_1}}{L_k^{sk_2}}
\end{equation}

where $L_k^{sk_1}$ is the bone length between the $k$-th joint and its parent joint in $sk_1$, $L_k^{sk_2}$ is the bone length between the $k$-th joint and its parent joint in $sk_2$.

\textbf{Smooth Transitions.} Depending on the values of \( t_s \) and \( t_i \) (for example, when \( t_s = 0 \) and \( t_i \) is fixed), we align the processed data by the root node, which, although keeps the positions of the human body on the original trajectory, inevitably results in discontinuities between consecutive frames. To mitigate this issue, we employ Spherical Linear Interpolation (SLERP) to smooth the transitions. Specifically, for the frames before and after the \( t_i \)-th frame, we apply SLERP over five frames to interpolate between them. SLERP provides a smooth transition between two rotations by interpolating along the shortest path on a sphere, ensuring continuous motion. This interpolation technique allows us to create more fluid and natural motion sequences.

Additionally, when computing \( L_{\text{rec}} \), we mask the features corresponding to these frames to avoid noise caused by abrupt transitions, and apply \( L_{\text{smooth}} \) to handle this smoothing process. This ensures that the model's training is not influenced by noisy data, while also enabling it to learn the ability to transition smoothly between different motion segments. 

\textbf{Orientation Adjustments.} When inserting different data, although we ensure that the position of the root point remains relatively unchanged, errors in orientation may occur. Specifically, the orientation may either remain the same as the previous frame or be rotated 180 degrees around the z-axis. This issue arises because we use the 263-dimensional recursive representation of the HumanML3D dataset. During the recursion process, there can be singularities in the representation of the rotation angle, meaning that a rotation can represent orientation information that differs by 180 degrees. To address this, we add an extra cross-product check. For the last frame of the previous data and the first frame of the inserted data, we compute the cross product of the root node and the corresponding two nodes on the inner thigh. We then check whether the dot product of the resulting vectors is greater than 0. If the dot product is greater than 0, there is no issue. However, if it is less than 0, we apply a 180-degree rotation transformation around the z-axis to the inserted data to correct the orientation.

\begin{figure}[tp]
    \centering
     \includegraphics[width=\linewidth]{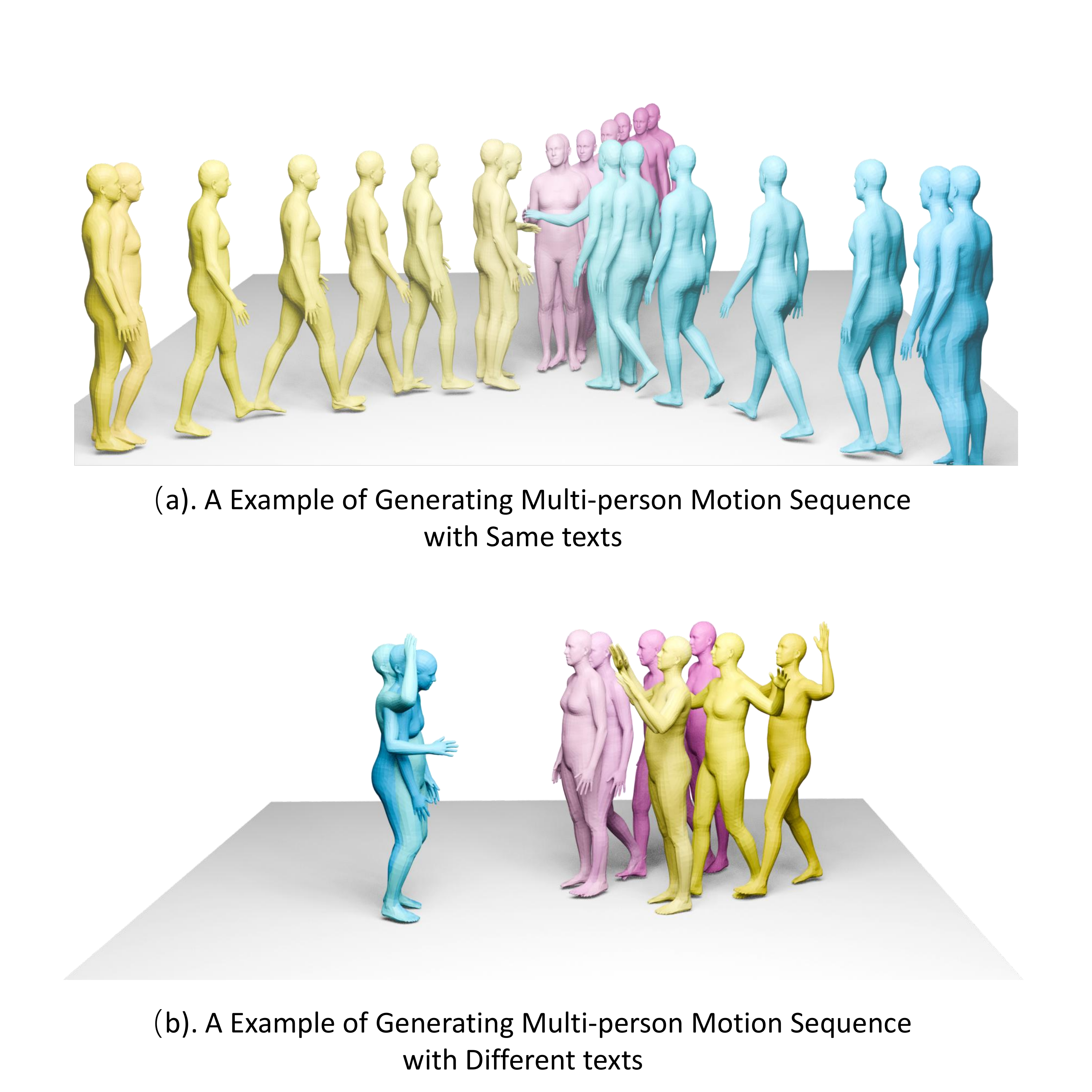}
    \caption{Multi-Person Motion Sequence Generation with Varied Text Inputs.}
    \label{multi}
\end{figure}

\subsection{Additional Experiments}
We conduct two sets of experiments to evaluate our model: the human study and the quantitative evaluation on HumanML3D dataset. In the human study, we evaluated models on text-motion alignment, motion naturalness, and dual-human interactivity, with 42 participants rating the results (1–5 scale). Our model consistently receives higher ratings across all criteria, demonstrating better semantic alignment and more natural, interactive motion generation. In the quantitative evaluation, our approach achieves higher retrieval accuracy, better generation quality, and improved multimodal consistency, confirming its effectiveness in producing accurate, diverse, and semantically aligned motions.
\begin{table*}[t]
\centering
	%\resizebox{0.8\columnwidth}{!}
    {
\begin{tabular}{lcccc}
\toprule
Human Study & Method & Text-Motion $\uparrow$ & Motion Naturalness $\uparrow$ & Interactivity $\uparrow$ \\
\cmidrule(lr){2-5}
 & FreeMotion &  $2.57$ & $3.41$  &  $2.17$ \\
 & Ours & $3.86$  & $4.02$ &  $3.64$  \\
\midrule
Random Samples & Method & R Precision Top 1 $\uparrow$ & R Precision Top 2 $\uparrow$ & FID $\downarrow$ \\
\cmidrule(lr){2-5}
 Single-person Dataset & FreeMotion &  $0.179^{\pm .022}$ & $0.294^{\pm .017}$  & $ 1.785^{\pm .068}$ \\
 & Ours & $\mathbf{0.204^{\pm .002}}$ & $\mathbf{0.357^{\pm .008}}$ & $\mathbf{0.740^{\pm .023}}$ \\
\cmidrule(lr){2-5}
 & Method & MM Dist $\downarrow$ & Diversity $\rightarrow$ & MModality $\uparrow$ \\
\cmidrule(lr){2-5}
 & FreeMotion &  $5.261^{\pm .009}$ & $9.372^{\pm .090}$ & $1.028^{\pm .059}$ \\
 & Ours & $\mathbf{3.928^{\pm .010}}$ & $9.127^{\pm .045}$ & $\mathbf{1.476^{\pm .021}}$\\
\bottomrule
\end{tabular}
}
\caption{Experimental results for human study and experiments in single-human dataset, respectively.}

\label{addition}
\end{table*}

\subsection{Multi-Agent Motion Generation Results and Analysis}
\begin{figure*}[t]
    \centering
    \includegraphics[width=\linewidth]{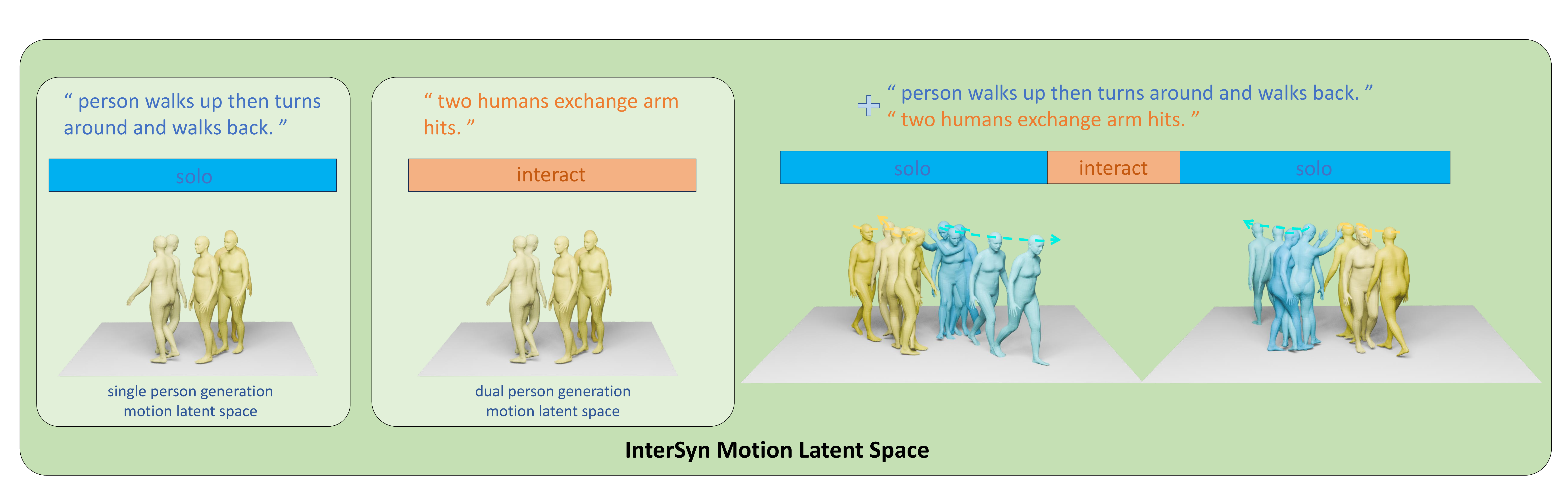}
    \caption{Additional Generation of Interleaved Motion Sequences.}
    \label{exp}
\end{figure*}
In our work, we first introduced the framework for generating single-agent motions and then extended it to handle dual-agent interactions through the concept of interleaved motion synthesis. This approach allows the model to generate motion sequences for two agents, where each agent's motion is interwoven in a way that respects both individual action and interaction between the agents. By leveraging the conditional information, we generate coherent sequences for each agent, ensuring that their motions are synchronized and contextually meaningful in the interaction. This dual-agent interaction framework serves as a foundation for extending the model to more complex scenarios.

Building upon this, our method is capable of generating motion sequences for single agents, dual-agent interactions according to Fig \ref{exp}, and can even be extended to multi-agent scenarios. This scalability is made possible by the modular design of our system, where the core principles of motion synthesis and coordination are applied iteratively across multiple agents. The model can generate multiple independent motion sequences and coordinate interactions across agents, ensuring that even in more crowded or complex scenes, the generated motions remain natural and coherent.

We can generalize the dual-person interaction alignment module to multiple agents by leveraging the properties of the coordinator in the main text. Specifically, we generate multiple independent motion sequences (including interactive actions) through the INS module. Each motion sequence can be treated as a sequence generated under the constraints of other motion sequences. We iteratively feed these motions into the coordinator, using the example of three-person motions. Figure \ref{multi} illustrates an example of three-person motion generation. Although our current network can support the generation of multi-agent motions, the model's sensitivity to interaction constraints is still limited. Additionally, it requires relatively structured and routine actions (e.g., walking over and shaking hands) to generate high-quality results. For more complex actions (e.g., A greets B, then hugs C), the model has limited generalization ability with respect to spatial positions. In future work, we may consider modeling trajectory constraints for multiple agents.
\begin{figure*}[t]
    \centering
    \includegraphics[width=\linewidth]{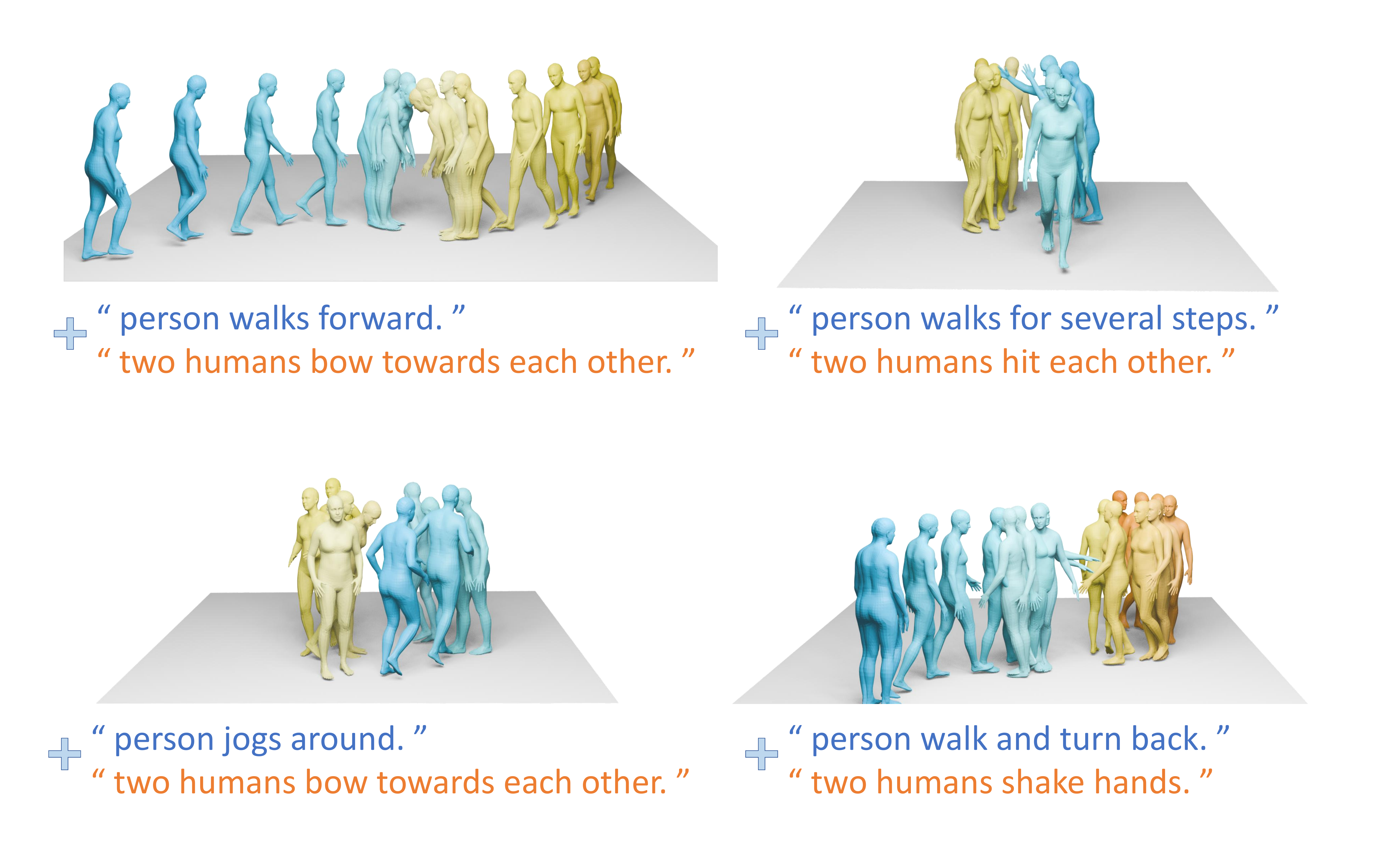}
    \caption{Additional Generation of Interleaved Motion Sequences.}
    \label{fig:method}
\end{figure*}

\subsection{Additional Visual Results}
In this section, we present additional visual results to further demonstrate the effectiveness of our proposed method. The provided visualization highlights the motion synthesis and coordination processes across different scenarios. By visualizing both the individual and coordinated actions, we can observe how the model handles complex interactions between characters, ensuring smooth transitions and realistic behavior. These results offer further insight into the model’s ability to synthesize diverse motion sequences and manage coordination tasks, such as maintaining proper alignment and synchronization across multiple agents in a dynamic environment.

% \ifarxiv \clearpage \appendix \input{12_appendix} \fi

\end{document}